\documentclass[10pt,twocolumn,letterpaper]{article}

\usepackage{cvpr}              
\usepackage[dvipsnames]{xcolor}

\definecolor{cvprblue}{rgb}{0.21,0.49,0.74}
\usepackage[pagebackref,colorlinks,citecolor=cvprblue]{hyperref}

\def\paperID{027} 
\def\confName{3DV\xspace}
\def\confYear{2025\xspace}

\title{Spatial Cognition from Egocentric Video:\\ Out of Sight, Not Out of Mind}

\author{Chiara Plizzari\textsuperscript{1}\thanks{Work carried during Chiara's research visit to the University of Bristol}
\quad
Shubham Goel\textsuperscript{2,3}
 \quad
Toby Perrett\textsuperscript{4}
\quad
Jacob Chalk\textsuperscript{4}\\
\quad
Angjoo Kanazawa\textsuperscript{3}
\quad Dima Damen\textsuperscript{4} \\
\vspace*{-8pt} 
\and
\textsuperscript{1}Politecnico di Torino$\quad$
\textsuperscript{2}Avataar Inc. $\quad$
\textsuperscript{3}University of California, Berkeley\\
\textsuperscript{4}University of Bristol \\
\small{\url{http://dimadamen.github.io/OSNOM}}
}

\usepackage{graphicx}
\usepackage{booktabs}
\makeatletter
\apptocmd{\@maketitle}{\centering\insertfig}{}{}
\makeatother
\newcommand{\board}{\textcolor[RGB]{180,180,0}}
\newcommand{\knife}{\textcolor[RGB]{250,0,0}}
\newcommand{\plate}{\textcolor[RGB]{0,180,0}}

\usepackage{wrapfig}
\usepackage[accsupp]{axessibility}  

\usepackage{graphicx}
\usepackage{etoolbox}

\newcommand{\insertfig}{
    \includegraphics[width=\linewidth, trim=0 12 0 39]{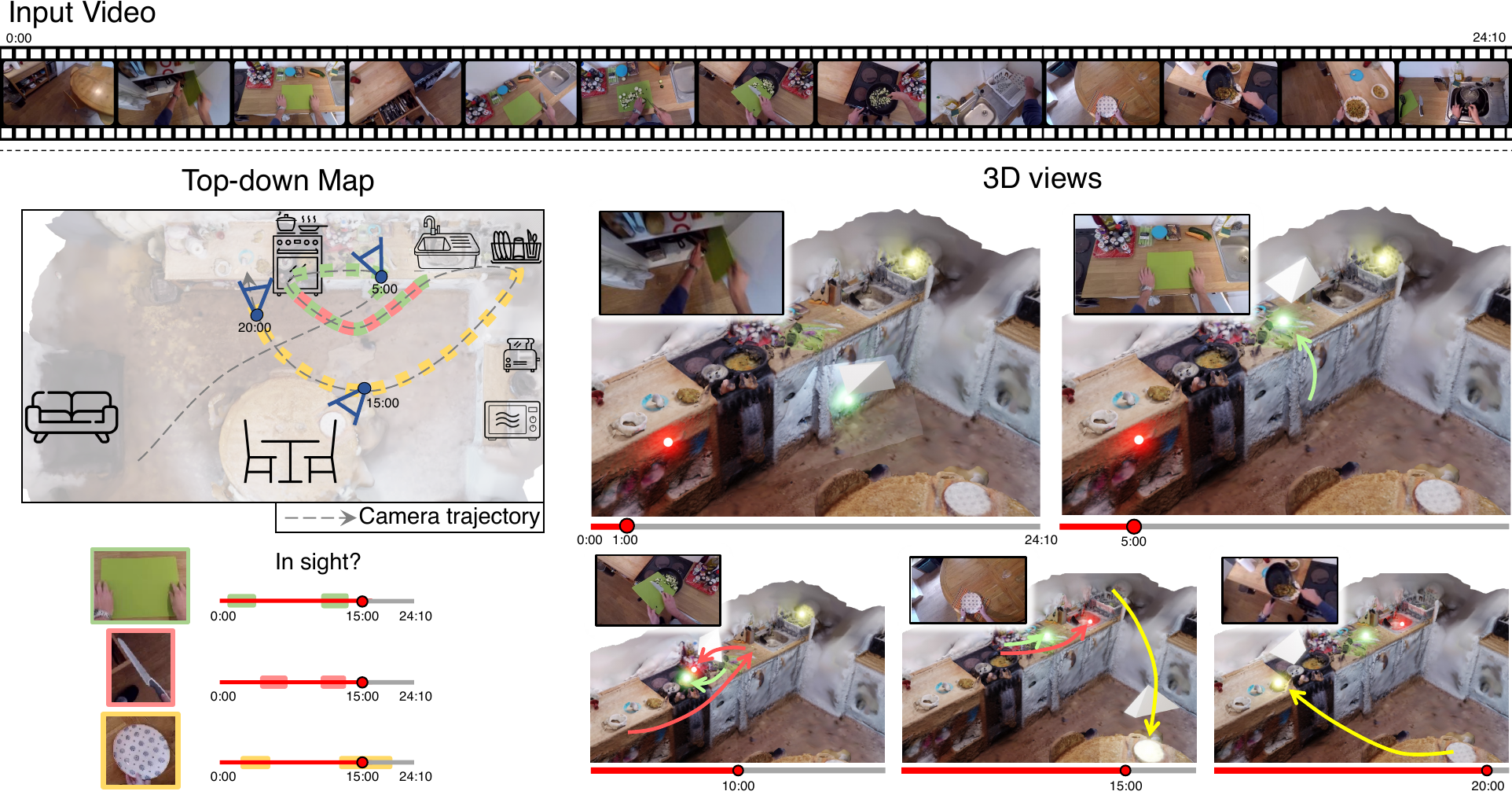} 
    \captionof{figure}{\label{fig:teaser}\small \textbf{Spatial Cognition.}
    From an egocentric video (top), we propose the task \textit{Out of Sight, Not Out of Mind}, where the 3D locations of all active objects are known when they are both in- and out-of-sight. 
    We show a 24 mins video along with world-coordinate tracks of 3 active objects through the video -- from a top-view down with camera motion (left top); identifying when they are in-sight (left bottom); their trajectory from a side view at five different frames (right). Neon balls show the 3D locations of these  objects over time along with the camera  (white prism), corresponding frame (inset) and object location change (coloured arrow).
    The \plate{chopping board} is picked from a lower cupboard (1:00) and is in-hand at 05:00. 
The \knife{knife} is picked up from the drawer (after 05:00), while in use (10:00) until it is discarded in the sink (before 15:00). The \board{plate} travels from the drainer to the table (15:00), then back to the counter (20:00). 
\vspace*{5pt}
    }
    
}
\makeatletter

\begin{document}
\maketitle

\begin{abstract}
\vspace*{-10pt}
As humans move around, performing their daily tasks, they are able to recall where they have positioned objects in their environment, even if these objects are currently out of their sight. In this paper, we aim to mimic this spatial cognition ability. We thus formulate the task of \textit{Out of Sight, Not Out of Mind} -- 3D tracking active
objects using observations captured through an egocentric camera. We introduce a 
simple but effective approach to address this challenging problem, called Lift, Match, and Keep (LMK). LMK~\textbf{lifts} partial 2D observations to 3D world coordinates, \textbf{matches} them over time using visual appearance, 3D location and interactions to form object tracks, and \textbf{keeps} these object tracks even when they go out-of-view of the camera.
We benchmark LMK on 100 long videos from EPIC-KITCHENS.
Our results demonstrate that spatial cognition is critical for correctly locating objects over short and long time scales. E.g., for one long egocentric video, we estimate the 3D location of 50 active objects. 
After 120 seconds, 57\% of the objects are correctly localised by LMK, compared to just 33\% by a recent 3D method for egocentric videos and 17\% by a general 2D tracking method.
\end{abstract}

\vspace*{-12pt}
\section{Introduction}
\label{sec:intro}

\textit{It is lunch time and the pan is on the hob. You bend to pick the chopping board
from a lower cupboard and put it on the counter. You then retrieve a knife from
the cutlery drawer. You use the chopping board and knife to slide chopped food
into the pan before discarding both in the sink. You then retrieve a clean plate
from the drainer to serve the food. As you move around the kitchen, you are
aware of where these objects are even if they are currently out of view.}

This ability to ``know what is where'' is an integral part of \textit{spatial cognition}. It allows humans to build a mental map of their environment, including ``memories of objects, once perceived as we moved about'' \cite{downs1973image}. Importantly, spatial cognition dictates these objects exist independently of human attention, and continue to exist in the cognitive map when the observer has left the vicinity \cite{moore2004object,committeri2004reference,burgess2006spatial,zewald2022object}. Spatial cognition is an innate ability, crucial to human survival, as it is how humans ``acquire and use knowledge about their environment to determine where they are, how to obtain resources, and how to find their way home''~\cite{waller2013handbook}.

In this paper, we operate on egocentric videos and make three prime \textbf{[C]}ontributions.\\ \textbf{[C1]} We introduce the task Out of Sight, Not Out of Mind~(OSNOM) -- maintaining the knowledge of where \textit{all objects} are, as they are moved about and even when absent from the egocentric video stream.
Egocentric views allow detailed observation of objects during interactions, e.g. the camera can look into the fridge or oven, and see exactly what was picked from the drainer. However, objects often swiftly move out of the camera's field of view.
We focus on these challenging set of \textit{active objects} that are moved by the camera wearer during the video sequence. Our task is to position \textbf{multiple} \textbf{dynamic} objects in 3D throughout the video, \textbf{both in- and out-of-view}.
This is distinct from existing tasks, such as episodic memory~\cite{grauman2022ego4d}, which search for the presence of an object within the video, i.e. within the camera's field of view.
Instead, the OSNOM task evaluates the locations of objects even when they are out of sight.
Figure~\ref{fig:teaser} illustrates the OSNOM task. 

To address the OSNOM challenge, \textbf{[C2]} we propose a simple but effective approach that tracks objects in the world coordinate frame. Specifically, we~\textit{lift} observations to 3D -- by reconstructing the scene mesh and projecting 2D detections given their depth from camera and surface estimates. We then \textit{match} these lifted observations using appearance and location over time to form consistent object tracks, and \textit{keep} the knowledge of objects in mind when they are out of sight.
This \textit{lift, match and keep (LMK)} approach allows spatio-temporal understanding which humans take for granted, yet is out of reach of current methods --- knowing when an object is within reach but is out-of-view, or when in-view but occluded inside a cupboard.

\textbf{[C3]} We benchmark OSNOM on 100 long videos from the EPIC-KITCHENS dataset~\cite{damen2020rescaling} through past and future 3D location estimations over multiple timescales.
We showcase that objects are out of view for $85\%$ of frames on average.
Using our LMK approach, we can correctly position $64\%$ of the objects successfully after 1 minute, $48\%$ after 5 minutes, and $37\%$ after 10 minutes, consistently outperforming recent approaches for ego~\cite{mai2022localizing,zhao2023instance} and general~\cite{zhang2022bytetrack} tracking.
Ablations demonstrate that maintaining 3D object locations over time is critical for correctly locating moving objects, and when they are occluded or out of view.

\section{Related Works}
\label{sec:related}
\noindent\textbf{Egocentric vision} has traditionally focused on tasks within the recorded video stream, i.e. within the camera's field of view.
These include understanding actions, objects and interactions over short, and more recently longer \cite{damen2020rescaling,grauman2022ego4d,darkhalil2022epic,tang2023egotracks}, timescales. Even when addressing future prediction (\eg action anticipation~\cite{Girdhar_2021_ICCV}), memory (\eg episodic memory~\cite{grauman2022ego4d}), object tracking~\cite{tang2023egotracks}, approaches scan the video stream to find when an object is in-sight. 
The seminal work Ego-Topo~\cite{nagarajan2020ego} builds a 2D affordance graph of the environment, relating actions to automatically discovered hotspots.
The motivation to capture the relative location of an object to the camera wearer was explored in EgoEnv~\cite{nagarajan2022egocentric}, by pre-training on 3D simulated environments. 
It shows that such environmentally-informed representations can improve performance on down-stream tasks such as episodic memory.

A number of tasks have been recently proposed that require 3D understanding in egocentric vision,
 such as jointly recognising and localising actions in a 3D map~\cite{liu2022egocentric}.
In Visual Query localisation in 3D (VQ3D)~\cite{grauman2022ego4d}, given a query image of an object, the aim is to localise \emph{only one} 3D position -- when \emph{the object was last seen unoccluded and in view.} 
Tracking is thus unnecessary (\eg SOTA on VQ3D, EgoLoc~\cite{mai2022localizing}, is based on retrieval  and notes most objects are stationary). 
Recently, 
Zhao~\etal~\cite{zhao2023instance} propose tracking a single object in the world-coordinate frame from RGB-D videos. 
They track objects in 2D, then lift these to 3D using the sensor's depth.
Another recent work~\cite{hao2024egodt} estimates camera pose from Dust3r~\cite{dust3r_cvpr24}, to predict positions of static objects like parts of the sofa. These are then tracked using rotational transforms to maintain 3D object consistency. Our approach is complementary in that we do not track objects that remain static but instead focus on tracking objects that the person has moved.
Importantly, both approaches only track objects when in view. Distinctly, the OSNOM task locates objects both when in- and out-of-view, offering a complete spatial cognition of dynamic objects.

\noindent\textbf{3D egocentric datasets} are now becoming available
\cite{ravi2023odin,grauman2022ego4d,pan2023aria,darkhalil2022epic,Grauman_2024_CVPR}. Examples include Ego4D~\cite{grauman2022ego4d}, which provides 3D scans and sparse camera poses for 13\% of the dataset, Ego-Exo4D~\cite{Grauman_2024_CVPR} which captures multiple first- and third-person views, and the Aria Digital Twin~\cite{pan2023aria}, which contains both camera poses and object segmentation masks for its two environments. 
EPIC-Fields~\cite{tschernezki2023epic} provides a pipeline to extract point clouds and dense camera poses from egocentric videos, and provides camera estimates for the EPIC-KITCHENS dataset~\cite{damen2020rescaling} across 45 kitchens. We use the pipeline from EPIC-Fields~\cite{tschernezki2023epic} paired with dense active object masks from VISOR~\cite{darkhalil2022epic}. 

Most related to 3D object tracking,~\cite{zhao2023instance} provides a small-scaled dataset with instance-level annotations in both 2D and 3D (4.5 hours, 250 objects in 10 environments).
Instead, we use a more diverse dataset collected in an unscripted manner, with our annotations covering 25 hours, more than 2K objects in 45 environments. 
Similarly Ego3DT~\cite{hao2024egodt} is evaluated on very short videos (less than 1 minute long).

\noindent\textbf{Object tracking through occlusion} has been investigated in 2D,  where maintaining object permanence, through heuristics \cite{huang2005tracking} (\eg constant velocity \cite{breitenstein2009robust}) or learning \cite{shamsian2020learning,tokmakov2021learning}, can track assignment when occluded objects reappear. However, these works do not track out of the field of view, and evaluate on short-term sequences
(\eg TCOW~\cite{van2023tracking} uses sequences of maximum length of 464 \emph{frames}).

\noindent\textbf{Autonomous-driving} 
typically maintains a map of the vehicle's surroundings~\cite{wong2020mapping} and tracks nearby vehicles, even when out of sight. However, whilst maintaining object locations through occlusion \cite{gilroy2019overcoming,ren2021safety}, tracks are deleted regularly as the vehicle only has to know about its vicinity. 

\noindent\textbf{Human tracking} has seen progress from 2D \cite{zhang2022bytetrack, bergmann2019tracking, meinhardt2022trackformer}, to 3D \cite{rajasegaran2021tracking}, to 3D with motion models \cite{khurana2021detecting,rajasegaran2022tracking, goel2023humans} which predict the location of occluded humans. Although these approaches use 3D for tracking, they usually do so in the camera coordinate frame. 
Recent works have explored simultaneous reconstruction of camera motion and human pose in 3D~\cite{ye2023decoupling,kocabas2023pace,yuan2022glamr}, with~\cite{sun2023trace,ye2023decoupling} evaluating this concept on human tracking. \cite{khirodkar2023egohumans} proposes a benchmark for tracking humans from multiple ego- and exo-centric cameras.

We present the first egocentric video work to track multiple objects in the world coordinate frame. Unlike humans, objects in egocentric videos do not move autonomously and frequently enter and exit the camera view. Our approach focuses on dynamic objects, tracking them in 3D space even when out of view, preserving object permanence.
We detail our approach next.

\section{Method - Lift, Match and Keep (LMK)}
\label{sec:method}

Our method operates on a single untrimmed egocentric video, $E$, recorded in an indoor environment.
We aim to keep track of all objects of interest in the 3D world coordinate frame. These 3D tracks capture the locations of objects throughout the video, even when they are not visible in the camera frame, solving the task of Out of Sight, Not Out of Mind (OSNOM).

As many objects in the scene remain in the same position throughout the video, we focus on the challenging set of active objects that the camera wearer interacts with, typically moving these objects from one place to another, often multiple times in the video.

We take as input observations of active objects ${o_n = (f_n, m_n)}$, where $f_n$ is a frame, and $m_n$ is a semantic-free 2D mask in that frame given in \emph{image coordinates}. 
The set of all observations, across the whole video, is $\mathcal{O} = \{o_{n} : n=1, ..., N\}$. We call these observations \emph{partial}, as they do not exist for every object in every frame.
The number of observations $N$ is much larger than the number of active objects - each object may be the subject of multiple observations. $N$ is also independent of the number of frames $T$, as frames may contain zero or multiple masks.

We call our method Lift, Match and Keep~(LMK). We first \emph{lift} 2D observations of objects to 3D (Sec~\ref{sec:3d_obs}), \emph{match} them over time, and \emph{keep} objects in mind when they are out-of-sight (Sec~\ref{sec:spatial_cognition}).
We detail LMK next.

\begin{figure*}[t]
    \centering
    \includegraphics[width=\linewidth]{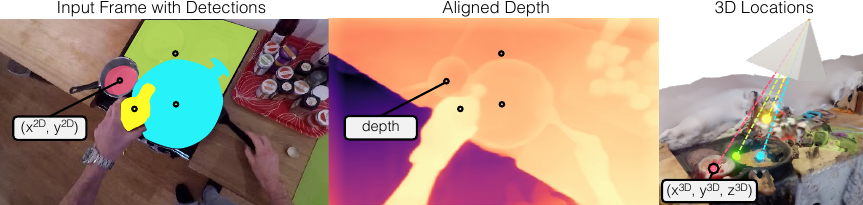}
    \caption{\textbf{Lifting 2D observations to 3D. }
    We use mask centroids as 2D object locations, sample corresponding depths from the mesh-aligned monocular depth estimate. We then compute the 3D object locations in world coordinates by un-projecting the mask's centroid from the estimated camera pose.
    }
    \label{fig:lift}
    \vspace{-15pt}
\end{figure*}

\subsection{Lift: Lifting 2D Observations to 3D}\label{sec:3d_obs}

\noindent\textbf{3D Scene Representation.}
Given a single egocentric video stream,  we follow the pipeline proposed in~\cite{tschernezki2023epic} to estimate camera poses and a sparse point cloud of the static scenes. 
We ignore redundant frames by calculating the homography over consecutive frames, thus allowing these long videos to be processed by Structure from Motion~(SfM) pipelines such as COLMAP~\cite{schoenberger2016sfm}.
The selected subset of video frames contains sufficient visual overlap to register all frames to the SfM point cloud and estimate a camera pose $C_t$ for every time $t$ in the video.
Note that the intrinsic parameters of the camera are also automatically estimated by this pipeline.

This reconstruction focuses on estimating the static background of the scene.
Objects in motion are deemed as outliers during matching and are accordingly ignored in the reconstructions.
The pipeline produces a sparse point cloud that cannot be used for positioning objects in 3D as it is missing the notion of surfaces. We convert these point clouds to surface representations as follows.

We extract scene geometry as a 3D mesh using a classical Multi-View Stereopsis pipeline \cite{furukawa2009accurate,schoenberger2016mvs} that runs patch matching to find dense correspondences between stereo image pairs, triangulates the correspondences to estimate depth, and fuses them together into a dense 3D point cloud with surface normals. We recover a scene mesh~$\mathcal{S}$ from the dense point cloud using Delaunay surface reconstruction \cite{cheng2013delaunay}. Examples of these meshes can be seen in Figure~\ref{fig:teaser}.

\noindent\textbf{Estimating 3D locations from monocular depth.}\label{sec:lift_3d}
For each frame, $f_n$, we estimate the monocular depth estimation using~\cite{yang2024depth}.
The advantage of using this approach is the ability to estimate the position of both static and dynamic objects, including objects that are in-hand.
However, this per-frame depth is incorrectly scaled and temporally inconsistent. We thus align it to the reconstructed 3D mesh --
via a scale-shift transformation that minimises the least squares error to the mesh's depth rendered from the estimated camera viewpoint. We refer to this as the \textit{aligned depth map}.

Given an observation $o_n=(f_n,m_n)$, we then assign a depth $d_n$ to observation $o_n$ corresponding to the centroid of the 2D mask $m_n$ on the aligned depth map. We take the object's 2D location in frame $f_n$, depth relative to the camera $d_n$, and camera pose ${C}_{f_n}$, and project the observation to the fixed 3D world coordinate, such that:
\begin{equation}
    [X_n, Y_n, Z_n]^T = \mathcal{C}_{f_n} \left [ {{d_n K^{-1} [x_n, y_n, 1]^T} \atop 1} \right ]
\end{equation}
where $K$ represents the camera's intrinsic parameters.
We denote this 3D location as $l_n\in \mathbb{R}^3$. We visualise \textit{lifting} to 3D in Figure~\ref{fig:lift}. 
Note that we represent each observation as a point in 3D following previous works \cite{mai2022localizing,grauman2022ego4d}.
These 3D observations are still partial and only on individual frames. 

\noindent\textbf{Visual features. }
In addition to the 3D location, we also compute visual features for each observation $o_n$ which we need to match observations over time into 3D tracks. We denote this as $v_n = \Psi(f_n, m_n)$, where $\Psi$ is a function that represents the visual feature extractor applied to the mask $m_n$ on the frame $f_n$.

\noindent\textbf{Lifted Visual Observations.} We incorporate the 3D locations and visual features to give our set of partial observations $\mathcal{W} = \{w_n : n=1, ..., N\}$ in the world coordinate frame, where $w_n = (f_n, l_n, v_n)$. 
We next describe how we match these observations over time to form 3D tracks.

\subsection{Match and Keep Lifted Observations}\label{sec:spatial_cognition}

Given the set of lifted observations, in this section we describe how to assign observations to consistent identities (\ie track objects) across time. Object permanence dictates that objects do not actually \textit{disappear} when occluded or are out of the egocentric camera's view -- humans use spatial cognition to maintain the knowledge of where objects are.

We process the egocentric video~$E$ online, mimicking human spatial cognition: an object's location is tracked only after it is first encountered and this is when it is kept in mind.

\noindent\textbf{Track definition.}
Each track $\mathcal{T}^j$ represents the set of observations belonging to the same object.
We refer to the set of all tracks at time $t$ as $\mathcal{T}_t$.
A track has one 3D location at each time $t$, whether the object is in-sight or not, and we refer to 
this location by $L(\mathcal{T}_t^j)$.

Additionally, the track has an evolving appearance representation over time.
It is calculated at time $t$ using the visual appearance of the most recent $\gamma$ visual features assigned to the track.
Averaging visual features enhances representation robustness. Limiting the average to $\gamma$ recent frames accounts for appearance changes over time (\eg a bowl may be full, dirty, then clean). The track's appearance at time $t$ is denoted as $V(\mathcal{T}_t^j)$.

\noindent\textbf{Track initialisation.}
If an observation $w_n$ represents a new, previously unseen object, i.e., is not matched to another track using the online matching described next, we initialise a new object track with this observation.
We define an initialisation function $\mathcal{I}$, which initialises a new $\mathcal{T}^{J+1}$, where $J$ tracks already exist, to the current 3D location and appearance of the observation $w_n$.
As this is the first observation of the object, the track is projected back in time from the start of the video. $\forall t \le f_n$:
\vspace{-2pt}
\begin{equation}
    \hspace{-2pt}\mathcal{I}(w_n) \rightarrow \mathcal{T}^{J+1} :
    L(\mathcal{T}^{J+1}_{t}) = l_n \text{ and } V(\mathcal{T}^{J+1}_{t}) = v_n
    \vspace{-2pt}
\end{equation}
This reflects the common sense that objects do not magically appear out of thin air, so the first encounter of an object is an indication of its presence in that location earlier.

\noindent\textbf{Track update.}
Once a track is initialized, its appearance and location are updated using new observations when available. The track update function $\mathcal{U}$ takes the track, observation and time as input:
\begin{equation}
    \hspace{-3pt}\mathcal{U}(\mathcal{T}^j, w_n, t) \rightarrow L(\mathcal{T}_t^j) = l_n \text{ and } V(\mathcal{T}_t^j) = \mu(v_n, \mathcal{T}^j)
\vspace{-1pt}
\end{equation}
where $\mu$ calculates the mean of the past $\gamma$ observations assigned to the track $\mathcal{T}^{j}$. 
If the track $\mathcal{T}^j$ is not assigned a new 
observation 
at time $t$ then its representation remains unchanged:
    $\mathcal{U}(\mathcal{T}^j, \varnothing, t) \rightarrow \mathcal{T}^j_t = \mathcal{T}^j_{(t-1)}$.

\noindent\textbf{Online Matching.}
We describe the process of forming tracks from online observations. 
We find the set of new observations at each $t$;
$\mathcal{W}_t = \{w_n \quad \forall n: f_n = t\}$.
Note that $\mathcal{W}_t$ is empty if there are no observations at time $t$. 

Given the first frame with at least one observation,
we initialise one track for each of these
   ${\mathcal{T}_t = \{ \mathcal{I}(w_n) \quad \forall w_n \in \mathcal{W}_t \}}$.
We next iterate over time and compare $\mathcal{W}_t$ to the set of trajectories at time $t-1$. Matching is based on a cost function using a combination of 3D distance and visual similarity, as in~\cite{rajasegaran2022tracking}. We model 3D similarity $\mathcal{\sigma}_{L}$ between an observation $w_n$ and a track $\mathcal{T}^j$ by an exponential distribution, and visual similarity $\mathcal{\sigma}_V$ by a Cauchy distribution:
\vspace{-2pt}
\begin{equation}\label{eq:5}
    \mathcal{\sigma}_{L}(w_n, \mathcal{T}^j) =  \frac{1}{\beta_{L}}\exp \left[-D(L(\mathcal{T}^j_{t-1}), l_n)\right]
\vspace{-2pt}
\end{equation}
\begin{equation}\label{eq:6}
    \mathcal{\sigma}_V(w_n, \mathcal{T}^j) = \frac{1}{1 + \beta_V D(V(\mathcal{T}^j_{t-1}), v_n)^2} 
\end{equation}
where $D$ is the Euclidean distance and $\beta_{L}$ and $\beta_{V}$ are relative weights for location and visual similarities.

We define the cost $\Phi$ of assigning an observation with an existing track as a combination of 3D and visual distance:
\begin{equation}
   \hspace{-3pt} \Phi(w_n,\mathcal{T}^j) =  -\log{(\mathcal{\sigma}_{L}(w_n, \mathcal{T}^j))}  -\log{( \mathcal{\sigma}_V(w_n, \mathcal{T}^j))}
\end{equation}
We then use a simple Hungarian algorithm as a robust method for associations as in~\cite{zhang2021fairmot,zhang2022bytetrack,sun2020transtrack}. Our matching algorithm $\xi$ computes $\Phi$ between every observation in $\mathcal{W}_t$ and the tracks $\mathcal{T}_{(t-1)}$\footnote{A threshold for assignment cost is set to $\alpha$}.
It returns a set of track assignments $A_t$ for time $t$, where $A_t^j = w_n$ indicates that the track $\mathcal{T}^j$ is to be assigned the observation $w_n \in \mathcal{W}_t$:
\vspace{-6pt}
\begin{equation}
    A_t = \xi(\Phi, \mathcal{W}_t, \mathcal{T}_{t-1})
\vspace{-2pt}
\end{equation}
We update tracks and initialise new tracks, such that:
\vspace{-2pt}
\begin{equation}
    \mathcal{T}^t \leftarrow
    \begin{cases}
    \mathcal{U}(\mathcal{T}^j, A_t^j, t) \quad \forall j\\
        \mathcal{I}(w_n) \quad \forall w_n \in \mathcal{W}_t : \left( \nexists j :  A_t^j = w_n \right)
    \end{cases}
\vspace{-2pt}
\end{equation} 
By following the proposed online matching, we have an estimate of the 3D location for every object for which there is at least one observation.

\subsection {LMK for object visibility and positioning} \label{sec:lmk_states}
\label{subsection:spatial cognition}

As a result of the spatial cognition enabled by the Lift-Match-and-Keep process, we are able to provide further information about the visibility of each object in relation to the camera wearer at time $t$. An object $j$ can be \emph{one} of:
\begin{itemize}
    \item \textbf{In-sight}: if the corresponding track is assigned an observation at time $t$, \ie $A_t^j \ne \varnothing$
    \item \textbf{Occluded}: if $L(\mathcal{T}_t^j)$ is within the field of view of the estimated camera ${C}_t$, but there is no corresponding observation  ($A_t^j = \varnothing$). Note that without additional knowledge we cannot distinguish between missing observations and occlusion.
    \item \textbf{Out-of-view}: if $L(\mathcal{T}_t^j)$ is outside the field of view of the camera ${C}_t$.
\end{itemize}
An object may also be referred to as \textbf{Out-of-sight} if it is either out-of-view or occluded (\ie in the camera's viewing direction but cannot be detected as it is behind or inside another object).

LMK also discloses the relative distance between the object and the camera-wearer or the static environment:
\begin{itemize}
    \item \textbf{In-reach}: if the distance from object $j$ to the camera's position at time $t$ is within the camera wearer's near space $\eta$: $D(L(\mathcal{T}_t^j), {C}_t) \le \eta$
    \item \textbf{Out-of-reach}: as in-reach, but if $D(L(\mathcal{T}_t^j), {C}_t) > \eta$.
    \item \textbf{Moved}: object $j$ has moved \emph{relative to the environment} between times $t_1$ and $t_2$ if $D(L(\mathcal{T}_{t_2}^j), L(\mathcal{T}_{t_1}^j)) \ge \epsilon$, where $\epsilon$ is a minimum threshold (to account for small errors in camera and object positions).
    \item \textbf{Stationary}: as moved, but $< \epsilon$.
\end{itemize}
Note that an object may be both occluded but in-reach.

\section{Experiments}

Section \ref{sec:benchmark}, introduces our benchmark for the OSNOM task. 
Section \ref{sec:exp_setup} details baseline methods.
Section \ref{sec:results} contains the main results and qualitative examples.
Section \ref{sec:ablation} ablates LMK, including its capabilities for spatial cognition. 

\subsection{Benchmarking OSNOM}\label{sec:benchmark}

\noindent\textbf{Dataset.}
We evaluate on long videos from the EPIC-KITCHENS~\cite{damen2020rescaling} dataset.
This dataset offers unscripted recordings of single participants --
all object motions are thus a result of the camera wearer moving objects around.
We select on 110 videos from EPIC-Kitchens, 12 minutes long on average, from 45 different kitchens, for a total of 25 hours of videos. 
We randomly select 10 validation videos for hyperparmeter tuning and 100 videos for evaluation.
These videos contain a total of 7.9M masks, which correspond to 
2939 objects.
We use the object semantic label only for calculating the ground truth for evaluation.

For most of our results, we use masks provided by VISOR~\cite{darkhalil2022epic}, which are interpolations of ground-truth masks. This allows us to assess LMK's performance without accumulating errors from a object detector. For completion, we also ablate these results with the usage of a semantic-free detector~\cite{Shan_2020_CVPR} in Section~\ref{sec:ablation}.

\begin{figure} 
\centering
\includegraphics[width=0.4\textwidth,trim=0 0 0 0]{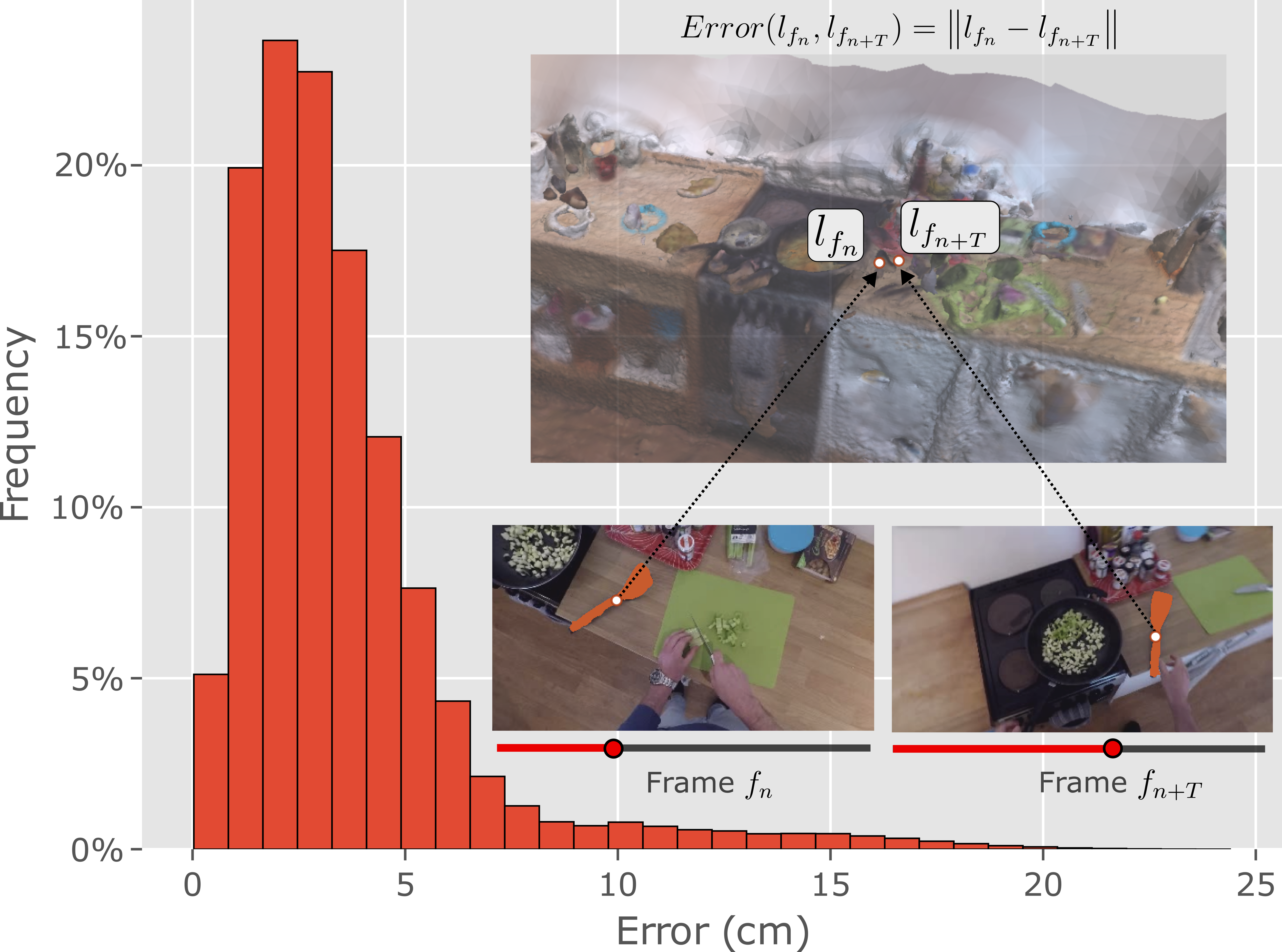} 
   \vspace{-6pt}
  \caption{\textbf{3D Projection error. }Distribution of Euclidean distance errors for the same object, at one location, comparing $l_n$ to $l_{n+T}$. }
  \label{fig:wrapfig}
  \vspace{-0.5cm}
\end{figure}

\noindent\textbf{Benchmark task. }\label{sec:eval}
Due to the length of our videos, an exhaustive evaluation for every object from every frame is intractable. 
We thus select challenging key-frames $\mathcal{F}$ -- these are frames with 3 or more objects being interacted with. 
Each frame $f \in \mathcal{F}$ includes objects that are in-sight and we wish to evaluate the methods' ability to correctly locate the 3D locations of these same objects over frames $f \pm \delta$.
We compare the performance of different methods as $\delta$ increases.
In total, we evaluate starting from $\mathcal{F} = 3299$ frames, locations at $603k$ frames and 2007 objects, averaging $49k$ frames and 20 objects per video.
Our benchmark is publicly available for comparisons (see Project Webpage).

\noindent\textbf{Ground truth locations.}
We use our 2D to 3D lifting approach presented in Section~\ref{sec:3d_obs} as ground-truth locations.
We quantitatively assess the error in these locations as follows.
We select a random set of objects and the corresponding time segments when these are in the same location throughout the environment. 
The error between the projections from multiple views, for the same object in the same location, allows assessing our 3D locations.
Our analysis (details in Appendix~\ref{sec:app:error}) shows that the mean 3D error is 3.5cm, with 88\% of all errors smaller than 6cm and 96\% of all errors smaller than 10cm (Figure \ref{fig:wrapfig}).
Given these results, we find our lifting to be sufficiently accurate to be used as ground-truth locations. This also informs our metric, where we ensure our threshold for accepting assignments is sufficiently larger than the error noted here.

\noindent\textbf{Evaluation metric.}
Traditional tracking metrics do not evaluate tracks when out of sight \cite{bernardin2008evaluating,ristani2016performance,luiten2021hota}. Thus, we define a metric called Percentage of Correct Locations (PCL), drawing inspiration from the Percentage of Correct Keypoints (PCK)~\cite{yang2012articulated} used to evaluate pose estimation, to evaluate the spatial alignment of objects. PCL considers a correct prediction at time $t$ if the object is correctly identified at time $t$ and its predicted 3D location is within a threshold $R$ from the ground truth 3D location. 
As PCL is calculated throughout time, any lost tracks are captured in the metric.

For our main experiments, we use $R=30\mathrm{cm}$\footnote{Half the standard width of a cupboard/cabinet which is 60cm/24inch}. This reflects that a function of spatial cognition is to know the location of an object with sufficient precision in order to navigate to or obtain it \cite{downs1973image,waller2013handbook}.
$R$ is visualised and ablated.

\label{sec:exp}

\subsection{Experimental setup}\label{sec:exp_setup}

\noindent\textbf{Baselines.}\label{sec:baselines}
As no prior works have attempted the OSNOM task, we compare LMK against three naive baselines and three previous works adapted to the OSNOM task: 

\begin{itemize}
     \item \textbf{Random Matching}: each observation is randomly assigned either to an existing track or a new track, demonstrating the complexity of the data. 
    \item Out of Sight, Lost \textbf{(OSL)}: objects are forgotten when they go out-of-view, so PCL is reported as 0 and their tracks are terminated. This baseline highlights the challenge in egocentric video, where objects move very frequently out of view soon after being first observed.    
    \item Out of sight, out of mind \textbf{(OSOM)}: observations can only be assigned to tracks which are in-view. When a track goes out-of-view, PCL is reported as 0 and tracks are frozen until it is back in-view.
    This is an upper bound for tracking in the camera coordinate frame.
    \item \textbf{ByteTrack} \cite{zhang2022bytetrack}: a strong, recent 2D multi-object tracking method, widely used as a baseline~\cite{zeng2022motr,Li_2024_CVPR,Qin_2024_CVPR}. Objects are tracked in 2D and then lifted in 3D using our lifting approach for evaluation.
    \item \textbf{EgoLoc} \cite{mai2022localizing}: we adapt this SOTA VQ3D approach to OSNOM, to handle multiple objects. We use the same masks and lifting for fair comparison. 
    EgoLoc's weighted averaging over all past observations fails for OSNOM because objects change position, so instead we take the most {recent} match. 
    \item \textbf{IT3DEgo} \cite{zhao2023instance}:
    As this paper uses ground truth depth which is not available in our RGB sequences, we instead run the 2D tracking using their public code, then lift the tracked objects to 3D using our approach (Sec~\ref{sec:lift_3d}).

\end{itemize}

\begin{figure}[t!]
        \centering
        \includegraphics[width=\linewidth, trim=0 8 0 0]{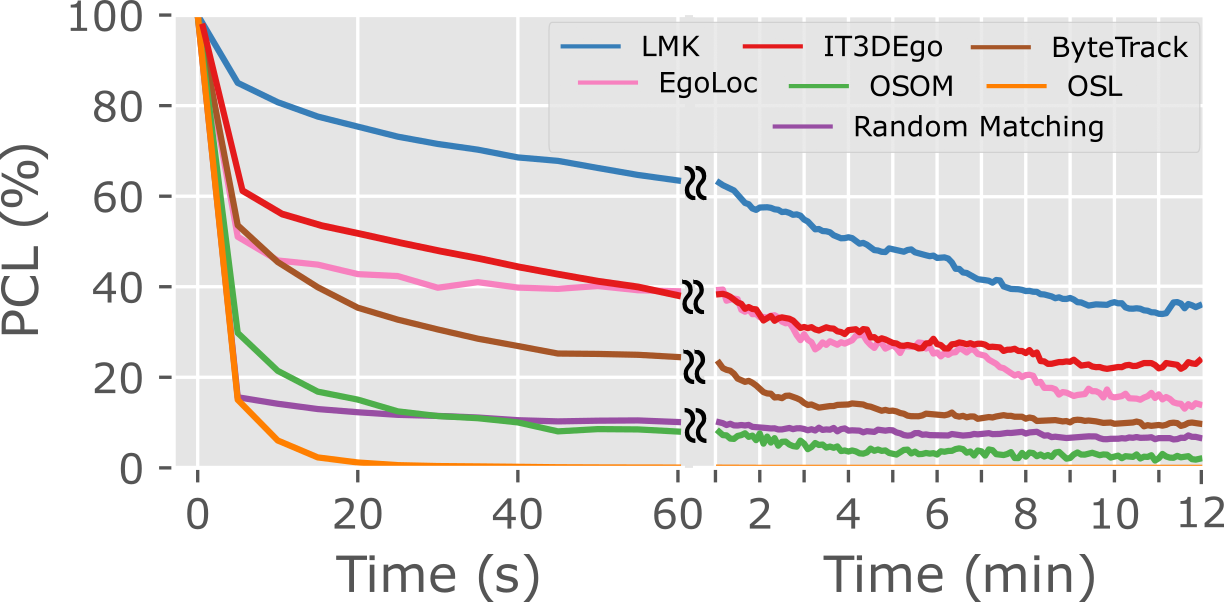}
        \caption{\textbf{OSNOM results. }PCL of LMK compared to baselines. }        \label{fig:spatial_cognition}
             \vspace{-10pt}
\end{figure}

\noindent\textbf{Implementation details. } For appearance features $\Psi$, we use a DINO-v2 \cite{oquab2023dinov2}. 
We crop each mask, scale to $224\times224$ and pass to the backbone.
We set $\alpha=10$, $\gamma=100$, $\beta_{L}=13$ and $\beta_{V}=2$ (chosen on the validation set). 
We compute meshes in advance, which takes 5 hours on average on one 2080Ti per video.
Then for online tracking,
DINOv2 operates at 
30 FPS and lifting-to-3D at 200FPS on one P100. LMK runs at 1000fps on a single CPU core.

\subsection{Results}\label{sec:results}

Results on the OSNOM task for LMK, compared to the baselines, are shown in Figure \ref{fig:spatial_cognition}. The average PCL (y-axis) over the whole dataset is reported for each 5s evaluation interval (shown on the x-axis), with standard deviation shaded. We show performance over both short (0-60 seconds) and long (1-12 minutes) timescales. Over time, the complexity of matching observations increases as more objects are being interacted with and tracked. This is reflected in performance decreasing for all methods over time.

LMK presents a significant improvement over all baselines.
The rapid drop in performance in OSOM and OSL shows the challenge of egocentric footage, where the constantly moving person causes objects to go out of view frequently. 
When objects are tracked as long as in-view (OSL baseline), performance goes to zero just after 20s, showing that objects are quickly lost from sight. The OSOM baseline shows that only considering objects in-view, without 3D world coordinates and object permanence, is insufficient for the OSNOM task (is worse than random).
LMK significantly outperforms ByteTrack, EgoLoc, and IT3DEgo. ByteTrack and IT3DEgo rely on 2D frames, while EgoLoc loses tracking quickly by comparing to initial appearances. In contrast, LMK tracks across consecutive frames, handling appearance changes from orientation or occlusion and leveraging 3D locations for robust matching.

\subsection{LMK Ablation}\label{sec:ablation}
\begin{figure}[t!]
        \centering
        \includegraphics[width=\columnwidth, trim= 0 15 0 0]{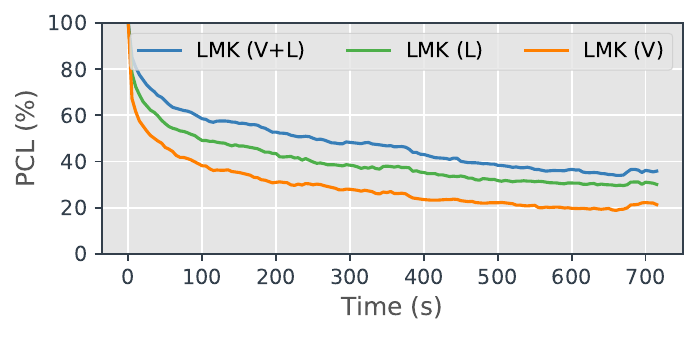}
    \caption{\textbf{Effect of visual appearance and location}. PCL for visual features (V), location features (L), or both (V+L).}
    \label{fig:location}
         \vspace{-3pt}
\end{figure}

\noindent\textbf{Effect of visual appearance and location.} 
LMK assigns observations to tracks based on 
appearance and location similarity. Figure \ref{fig:location} shows the effect of only visual appearance (V) 
and only location (L) 
compared to the default of both (V+L). Their combination shows improvements (mean +19\% over V, +8\% over L), highlighting that appearance and location are complementary. Appearance is good for frame-to-frame assignment, and location is particularly helpful for objects in motion, occluded and for reassigning objects when they reappear.

\begin{figure}[t!]
\vspace{-0.2cm}
\centering
\includegraphics[width=0.96\linewidth, trim= 0 15 0 0]{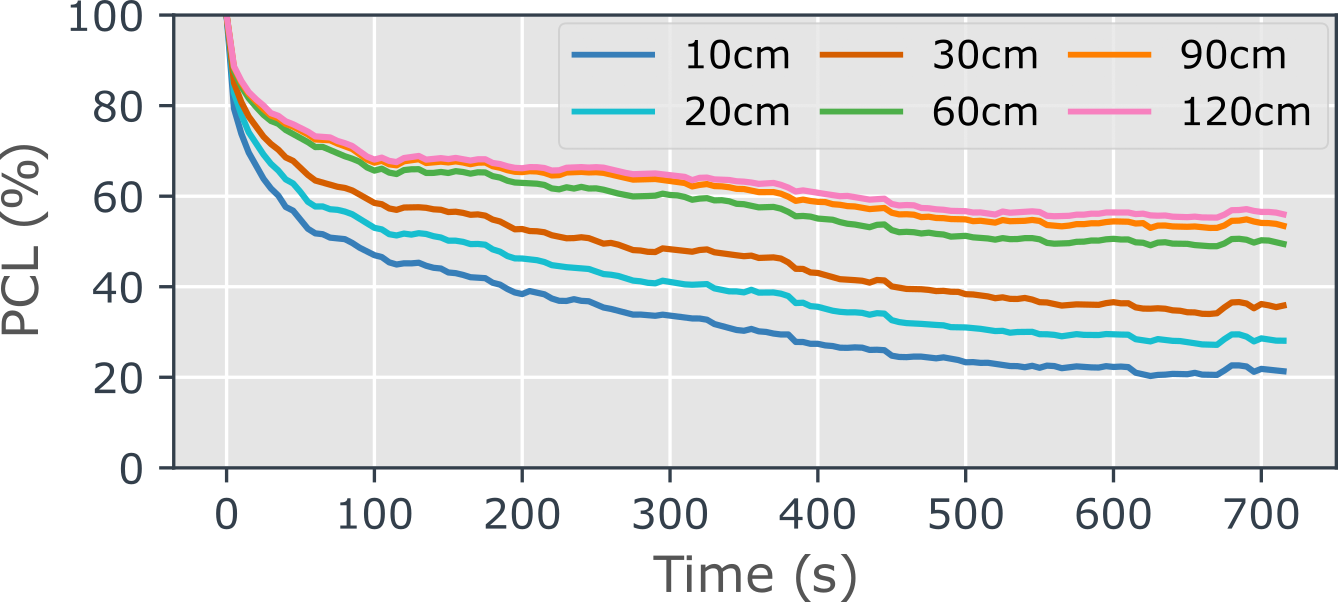}
        \caption{\textbf{Evaluation thresholds.} LMK when increasing the PCL threshold $R$ - the maximum distance between predicted and ground truth 3D locations considered successful. }
        \label{fig:r}  
        \vspace{-6pt}
\end{figure}
\begin{figure*}[t!]
     \centering
     \begin{minipage}[t]{0.33\textwidth}
         \centering
         \vspace{-92pt}
         \includegraphics[width=1\linewidth,trim= 5 0 4 0]{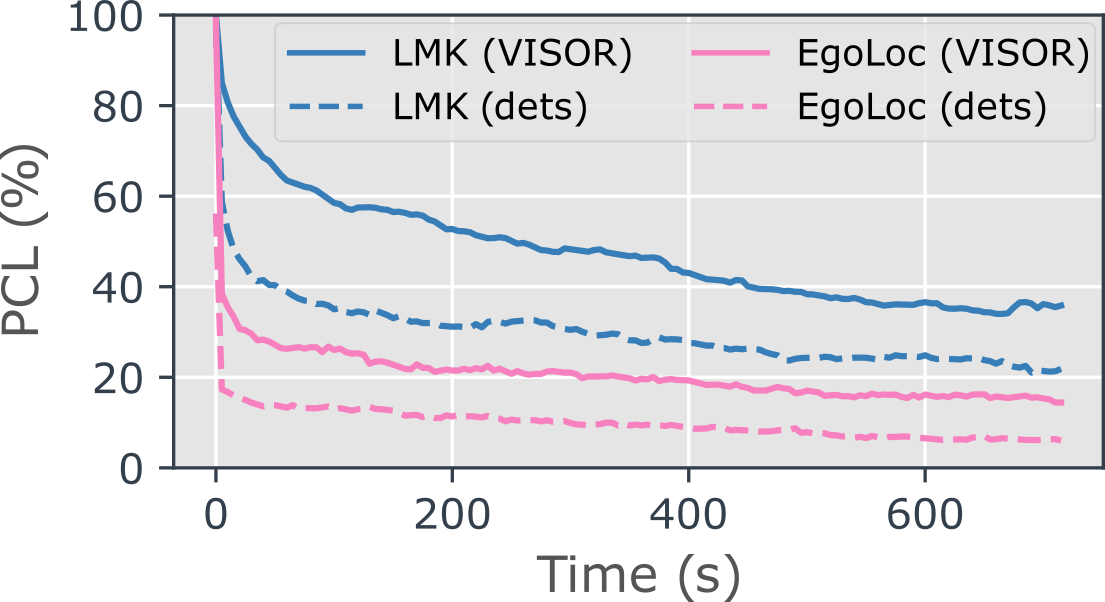}
        \caption{\textbf{Detections.} LMK and EgoLoc baseline~\cite{mai2022localizing} on both visual and location features (V+L) when using VISOR annotations \textit{vs} using object detections from \cite{Shan_2020_CVPR}. }
        \label{fig:dets}
     \end{minipage}
     \hfill%
     \begin{minipage}[t]{0.33\textwidth}
         \centering
         \includegraphics[width=\linewidth, trim= 0 10 0 0]{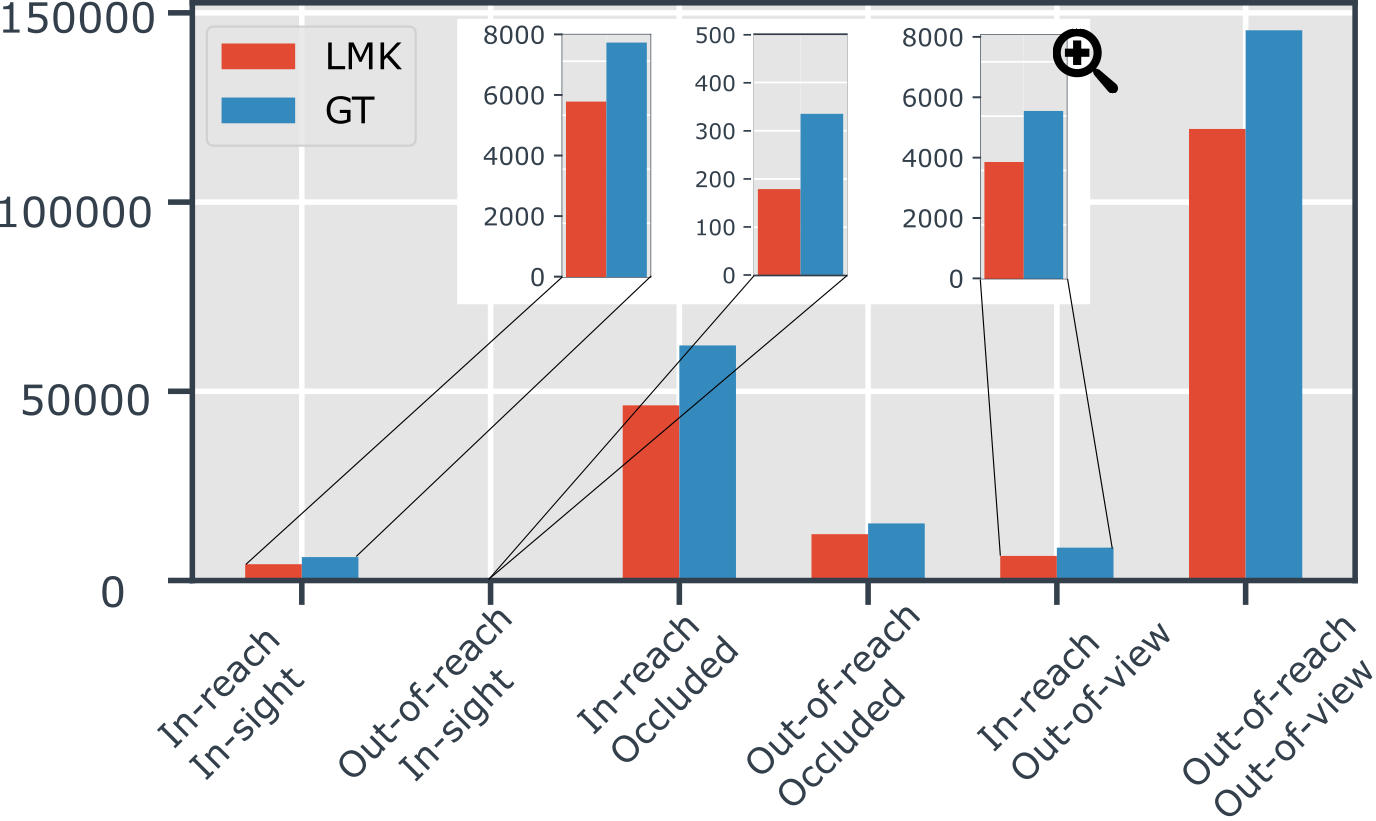}
         \caption{\textbf{LMK for spatial cognition. }Number of objects correctly located by LMK for each (In-reach, Out-of-reach) and (In-sight, Occluded, Out-of-view) combination.}
                  \label{fig:categories}
     \end{minipage}
     \hfill
     \begin{minipage}[t]{0.32\textwidth}
         \centering
         \includegraphics[width=\textwidth]{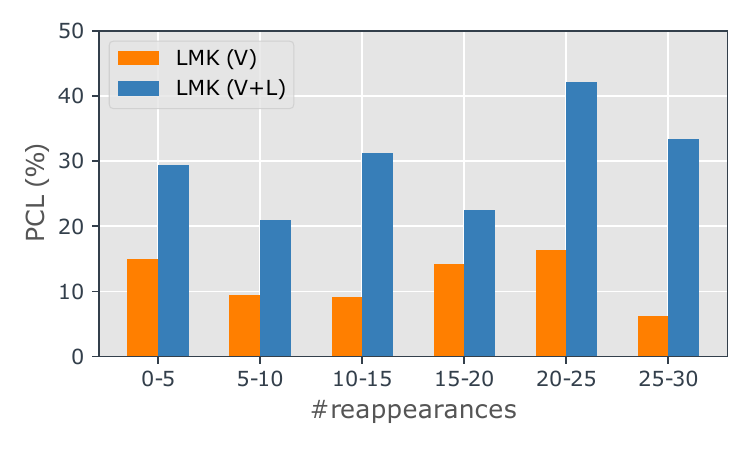}
    \caption{\textbf{Effect of reappearing.} Evaluation is performed over 10 minutes, for LMK with visual appearance (V) and the combination of visual appearance and location (V+L). }    \label{fig:reappear10}
         \end{minipage}
\end{figure*}

\begin{figure*}[th!]
    \centering
    \includegraphics[width=0.99\linewidth]{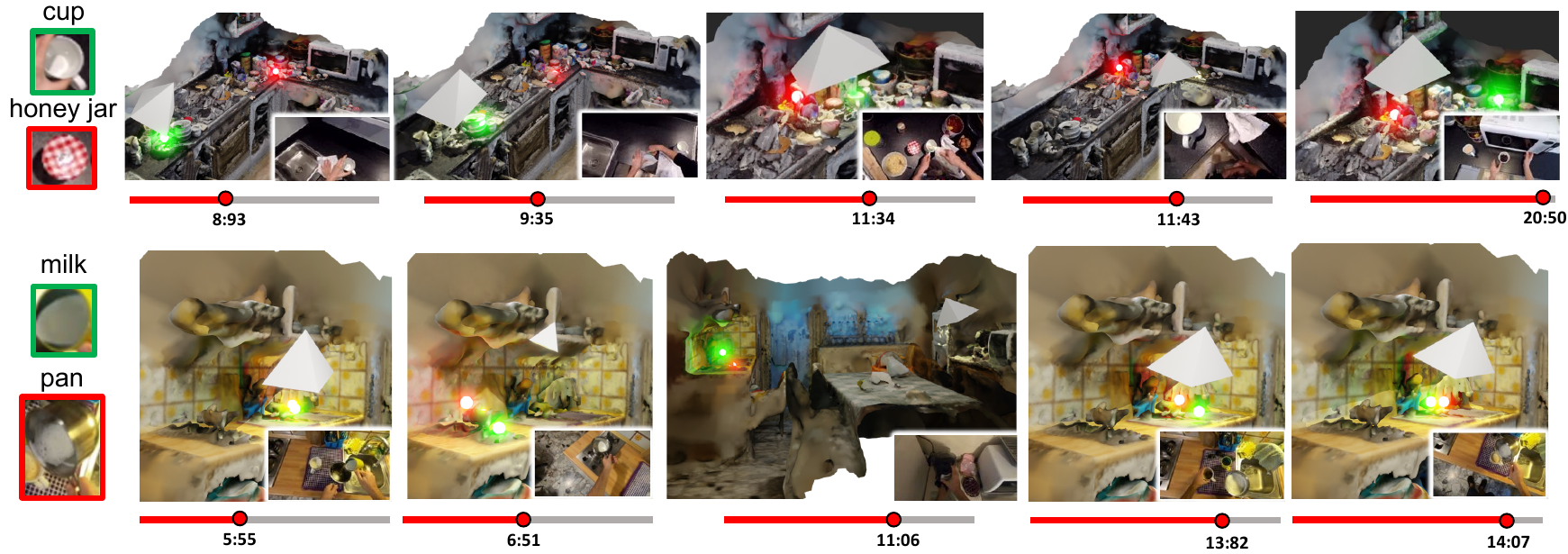}
    \vspace*{-10pt}
    \caption{\textbf{3D location prediction. }Predicted 3D locations (Neon dots) of two objects (left) over multiple times with frame insets (right). Note how object locations are accurately kept in mind, even when the camera-wearer is far away (bottom middle).}
    \label{fig:qual_2}
    \vspace{-10pt}
\end{figure*}

\begin{figure}[th!]
    \centering
    \includegraphics[width=\linewidth]{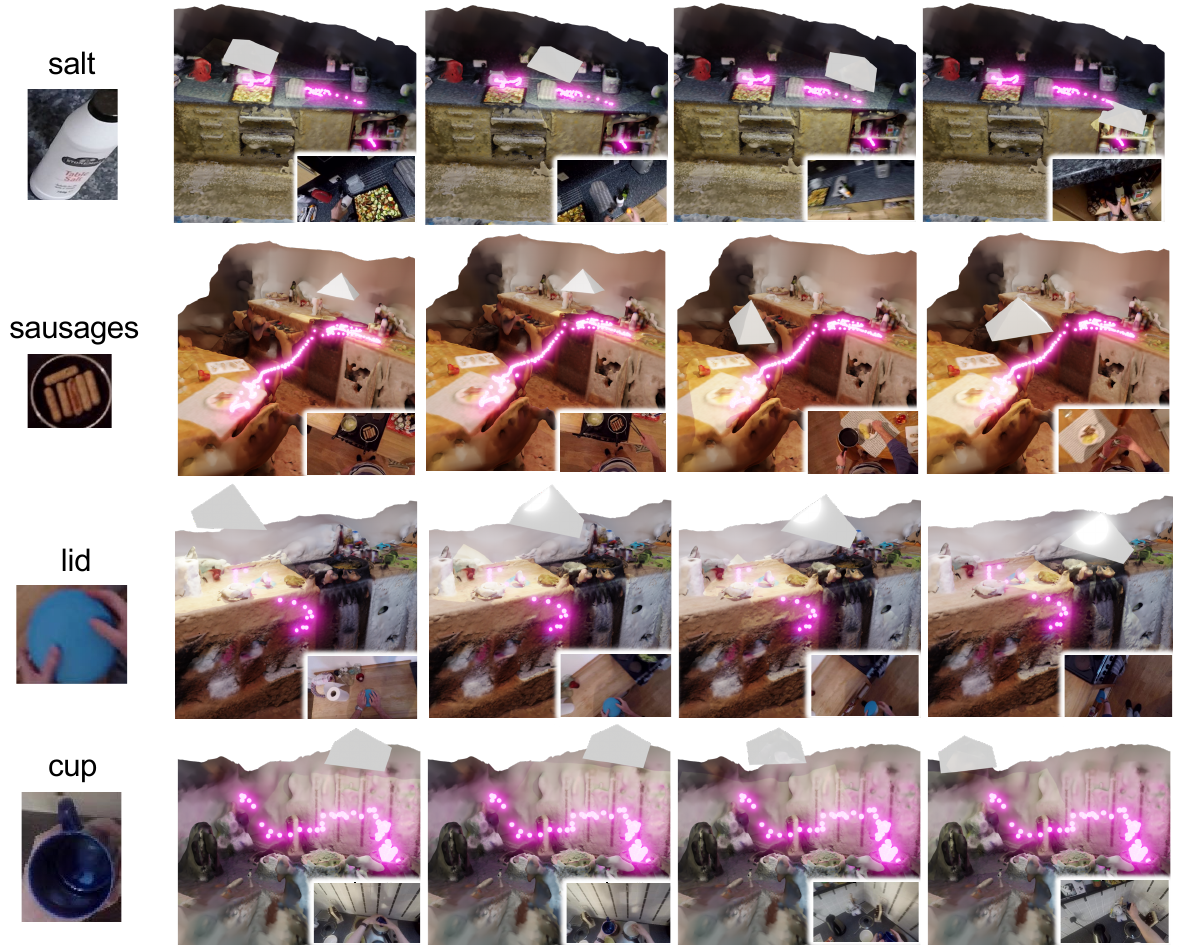}
    \caption{\textbf{Trajectory prediction} for objects in motion. Neon dots show correctly predicted 3D positions with corresponding camera views. Objects are accurately located both when static (on surfaces) and when moving (in-hand).} 
        \label{fig:qual_1}
    \vspace{-12pt}
\end{figure}

\noindent\textbf{Accuracy at different radii.} All our experiments set the PCL threshold, $R=30\mathrm{cm}$. Figure \ref{fig:r} also shows results when this is set to $R=10/20/60/90/120\mathrm{cm}$. As expected, PCL increases as $R$ increases.

\noindent \textbf{Detections.}  We used annotations from VISOR \cite{darkhalil2022epic} as 2D masks. This avoids compounding detector error when evaluating the error of 3D location estimation, which is our primary task.
In Figure \ref{fig:dets} we show an ablation using detections from \cite{Shan_2020_CVPR}. This model provides semantic-free bounding boxes of active objects, 
which we use as input to LMK and the best performing baseline EgoLoc. LMK still outperforms EgoLoc by a large margin.

\noindent\textbf{LMK for spatial cognition. }
Figure \ref{fig:categories} shows performance of LMK on the object states defined in Section \ref{subsection:spatial cognition}. For each combination of (In-reach\footnote{We use a reachable threshold $\eta=70\mathrm{cm}$}, Out-of-reach), (In-sight, Occluded, Out-of-view), we report the total number of ground truth objects and the number LMK correctly locates over a 1 minute interval. After 1 minute of objects being interacted with, LMK is still able to determine their locations, with an average accuracy of $72\%$. 
Additionally, LMK obtains 82\% on objects which are out-of-reach and out-of-view.

We also investigate the ability of LMK to track objects going out- then back in-view (\ie reappearing) within  10 minutes (Figure \ref{fig:reappear10}). 
LMK, matching using 3D locations, shows considerable performance improvement. 

\vspace*{12pt}
\noindent\textbf{\textbf{Qualitative results. }} 
Figure \ref{fig:qual_2} shows the predicted locations of a couple of objects at discrete time scales. %
In Figure \ref{fig:qual_1}, we show 3D trajectories of objects as they are moved around by the camera wearer. 
For example, we show the trajectory of the \textit{salt bottle} from being in the hand (pouring salt), placed on the countertop and eventually returned to a lower cupboard, while the \textit{cup} ends on a hanger. In all cases, 
LMK is capable of accurately tracking objects when when static (on surfaces) and when moving (in-hand).

We include examples of failure cases in Appendix~\ref{sec:failure}.

\section{Conclusion}

In this paper, we introduced the task of ``Out of Sight, Not Out of Mind'' (OSNOM) for egocentric video with partial object observations. It evaluates 3D tracking performance of active objects when they are both in- and out-of-sight. We introduced a very strong baseline: Lift, Match and Keep (LMK), a method which \emph{lifts} partial 2D observations in camera coordinates to 3D world coordinates, \emph{matches} them over time using visual appearance and 3D location, and \emph{keeps} them in mind when they go out of sight. Results on long videos from EPIC-Kitchens show LMK delivers good results over both short ($64\%$ after tracking for 1 minute) and long ($37\%$ after 10 minutes) timeframes, and that maintaining 3D world location is critical when objects go out-of-view. LMK outperforms recent works, strong 2D tracks and naive baselines by a big margin. For future work, we will investigate whether LMK can help track objects that undergo state changes, and explore shared 3D object tracks between multiple ego- and exo-centric cameras.

\noindent \textbf{Acknowledgments.} Research at Bristol is supported by EPSRC Fellowship UMPIRE (EP/T004991/1) \& PG Visual AI (EP/T028572/1).We particularly thank Jitendra Malik for early discussions and insights on this work.
We also thank members of the BAIR community and the MaVi group at Bristol for helpful discussions. This project acknowledges the use of University of Bristol’s Blue Crystal 4 (BC4) HPC facilities.

{
    \small
    \bibliographystyle{ieeenat_fullname}
    \bibliography{main}
}
\clearpage
\setcounter{page}{1}

\appendix

\section{Video examples}

Sample results are present on the project's webpage: \url{http://dimadamen.github.io/OSNOM}.
The video shows predicted object locations, over time, in 4 sampled clips from the evaluation EPIC-KITCHENS videos. We show the mesh of the environment, along with coloured neon dots representing the active objects that we lift and track in 3D.
The videos also show the estimated camera position and direction throughout the video along with the corresponding egocentric footage.

In each case, the clip shows object locations predicted when they are in-sight, when they are out-of-view as well as when they are moving in-hand. Selected examples also show objects picked up / returned to fridge or cupboard highlighting the complexity of spatial cognition from egocentric videos.

\section{Estimating error in the 3D projection} 
\label{sec:app:error}
In Section~\ref{sec:benchmark}, we estimate the error in 3D locations, through comparing projections of static objects from multiple viewpoint. 
Figure 3 in the paper presented the findings -- showcasing that the mean error is 3.5cm with 96\% of all errors within 10cm. We here describe the data used to report this figure.

We randomly selected 207,277 pairs of frames from our dataset, covering correspondences between 10 static objects across 5 different kitchens/environments.
These were manually selected to find multiple frames with masks of the same object, at distinct times, and from different viewpoints.
We avoid masks that are partially occluded by another object or by the camera's field-of-view (i.e. not fully in view) as these projections are likely to differ due to the occlusion of part of the mask.
As the chosen pairs of masks showcase the same static object, their 3D locations should perfectly match.
Any differences in their 3D location can be used to measure the error in the 3D projection, which we use as ground truth locations. 

As the figure showcases, the error in our projections is within 10cm and well-within the threshold we use of 30cm. Recall that our threshold is chosen to reflect the cupboard width in standard kitchens. Estimating an object's location within 30cm implies we can position the object correctly within a cupboard.

\begin{figure}[t]
        \centering
        \includegraphics[width=\linewidth, trim = 0 10 0 15]{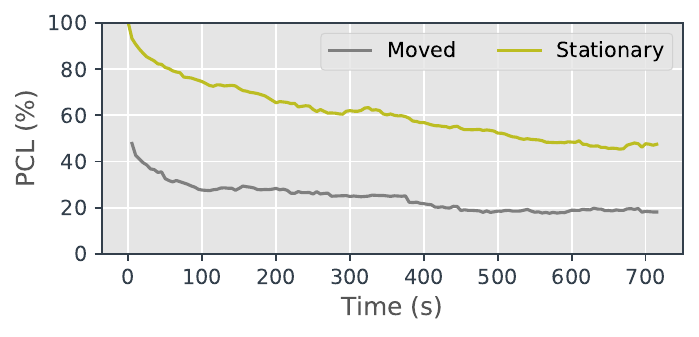}
        \vspace*{-3pt}
        \caption{LMK Results for \textbf{Moved vs Stationary} objects with respect to the environment. We used a movement threshold of $\epsilon=30\mathrm{cm}$}
        \label{fig:moved}
\end{figure}

\begin{figure}
    \includegraphics[width=\columnwidth, trim = 0 10 0 15]{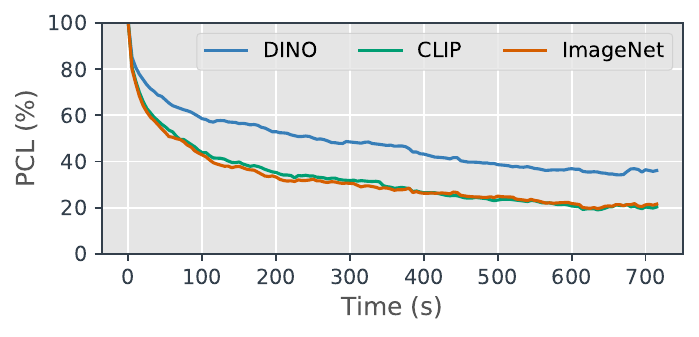}
    \caption{\textbf{Visual feature choice} of a DINO-v2, CLIP or ImageNet (ViT). }
    \label{fig:visual_feat}
\end{figure}%

\begin{figure}[t!]
     \centering
         \includegraphics[width=\linewidth, trim = 0 10 0 15]{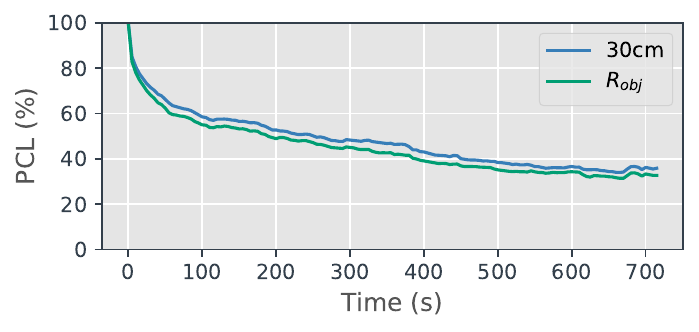}
   \caption{\textbf{Object radius.} LMK when approximating objects as spheres in 3D and using object radius for PCL threshold R.  }
         \label{fig:radius_objs}
\end{figure}

\begin{figure*}[h!]
\vspace{-5pt}
    \centering
    \begin{subfigure}{0.32\linewidth}
    \includegraphics[width=\columnwidth, trim = 0 10 0 15]{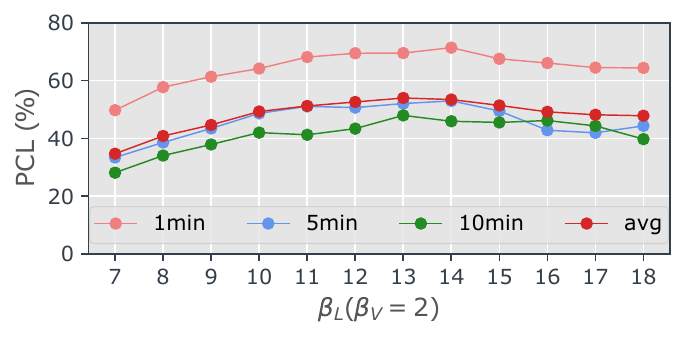}
    \caption{$\beta_L$, the weighting of 3D location for assigning observations to tracks.}
    \label{fig:beta_xyz}
    \vspace{-8pt}
\end{subfigure}
\hfill
\begin{subfigure}{0.32\linewidth}
    \includegraphics[width=\textwidth, trim = 0 10 0 15]{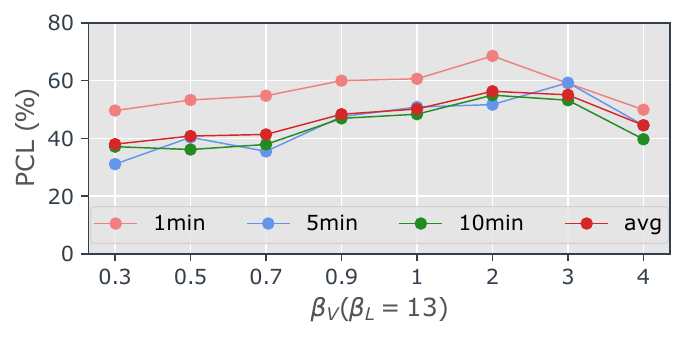}
    \caption{$\beta_V$, the weighting of visual appearance for assigning observations to tracks.}
    \label{fig:beta_v}
    \vspace{-8pt}
\end{subfigure}
\hfill
\begin{subfigure}{0.32\linewidth}
    \includegraphics[width=\textwidth, trim = 0 10 0 15]{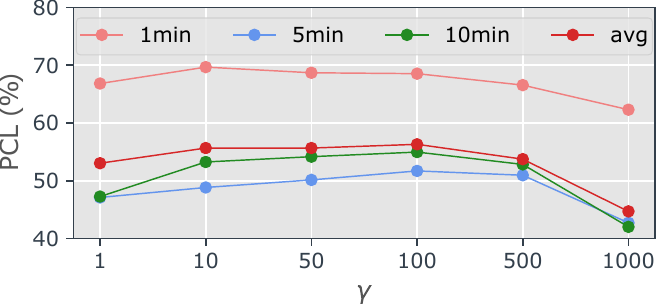}
    \caption{$\gamma$, the number of visual appearance features averaged to calculate track representation.}
    \label{fig:gamma}
    \vspace{-8pt}
\end{subfigure}
\caption{\textbf{Hyperparameter ablations} for LMK on the validation set. We choose the best average over 1, 5 and 10 minute sequence lengths.}
\vspace{-18pt}
\label{fig:enter-label}
\end{figure*}

\section{Additional Ablations}
\noindent\textbf{Moved \vs Stationary objects. } Section~\ref{sec:lmk_states} also provides a definition of objects which have either moved significantly within the environment or remained relatively within a small section of the environment. We use a movement threshold of $\epsilon=30\mathrm{cm}$ to separate large from small motions. Figure \ref{fig:moved} shows PCL results showing the objects that remain relatively stationary can be tracked on average $35\%$ better than that of objects which have moved significantly within the space. Objects are more visually different after a move (\eg different orientation or lighting).

\noindent\textbf{Visual features. } 
Our default feature extractor $\Phi$ is a ViT~\cite{dosovitskiy2020image}, pre-trained under the self-supervised DINO-v2 recipe \cite{oquab2023dinov2}. We also compare to ViTs pre-trained on CLIP \cite{radford2021learning} and ImageNet \cite{deng2009imagenet} in Figure \ref{fig:visual_feat}. DINO-v2 outperforms other approaches across all timescales, likely due to the pre-training tasks of CLIP (vision and language alignment) and ImageNet (image classification) being less suited to our requirement of reliable frame-to-frame visual similarity.

\noindent\textbf{Object size. } 
In our experiments, we use a fixed $R = 30cm$. As objects differ in size, one might argue that matching $R$ to the object size is more reasonable. In Figure \ref{fig:radius_objs} we use an adapted $R$ that matches the object dimension per example. Results are very similar to the default $R=30cm$, showcasing that fixed versus dynamic $R$ do not change the tracking capabilities.

\noindent\textbf{Weighting visual appearance and location.} LMK uses the hyperparameters $\beta_{L}$ (Eq~\ref{eq:5}) and $\beta_V$ (Eq~\ref{eq:6}) for relative weighting of visual and location similarities when assigning new observations to tracks. We select these based on best validation set performance averaged over timescales.
Figure \ref{fig:beta_xyz} shows validation set performance when fixing the chosen $\beta_V=2$ and varying $\beta_L$. Figure \ref{fig:beta_v} fixes $\beta_L=13$ and varies $\beta_V$. Both hyperparameters are relatively stable, most likely due to the scaling by appropriate distributions (Cauchy and Exponential).

\noindent\textbf{Track visual representation.}
Figure \ref{fig:gamma} ablates $\gamma$ over the validation set -- the number of recent features averaged for visual representation of a track. 
Best results are obtained with $\gamma=100$, with worse results for small / large values of $\gamma$, with performance relatively stable even down to only one observation.

\section{Failure cases}
\label{sec:failure}
\begin{figure*}[t!]
    \centering
    \includegraphics[width=0.9\linewidth]{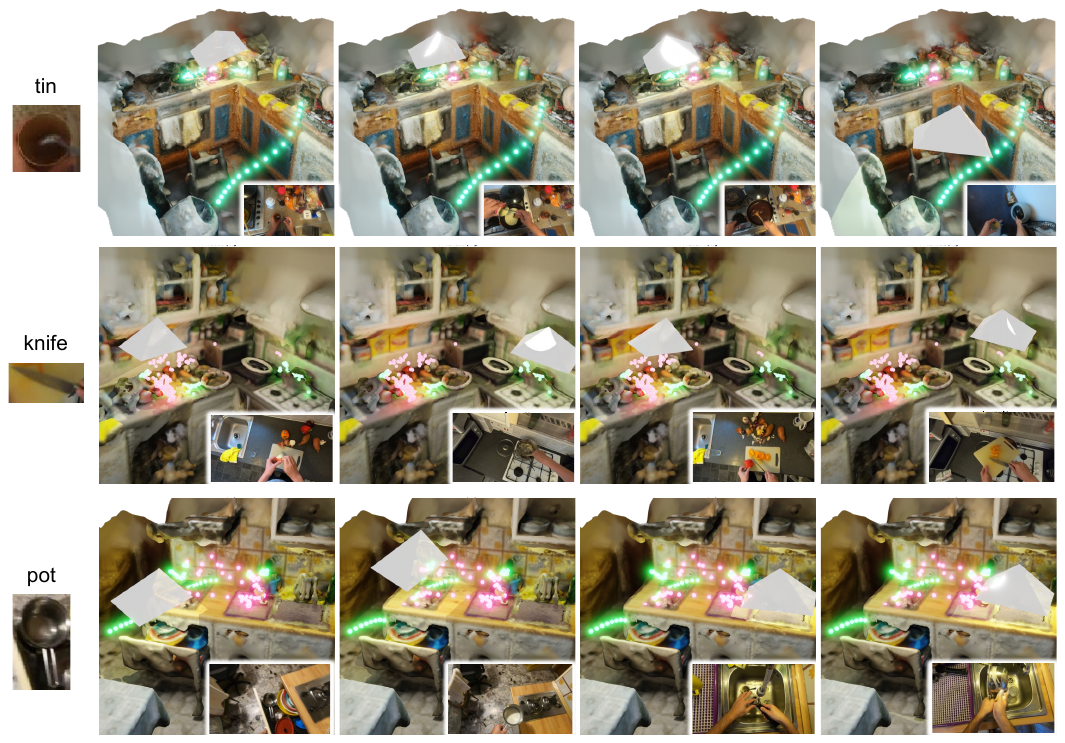}
    \caption{{\textbf{Trajectory prediction - temporarily lost but recovered track. }Predicted trajectory of three objects in motion. Green neon dots show correctly predicted 3D positions over four frames with their corresponding camera views, and red neon dots show ground-truth trajectory where the prediction fails. The tracking momentarily fails, but subsequently, the object is accurately matched to a future observation. }}
    \label{fig:failure1}
\end{figure*}

\begin{figure*}[t!]
    \centering
    \includegraphics[width=0.9\linewidth]{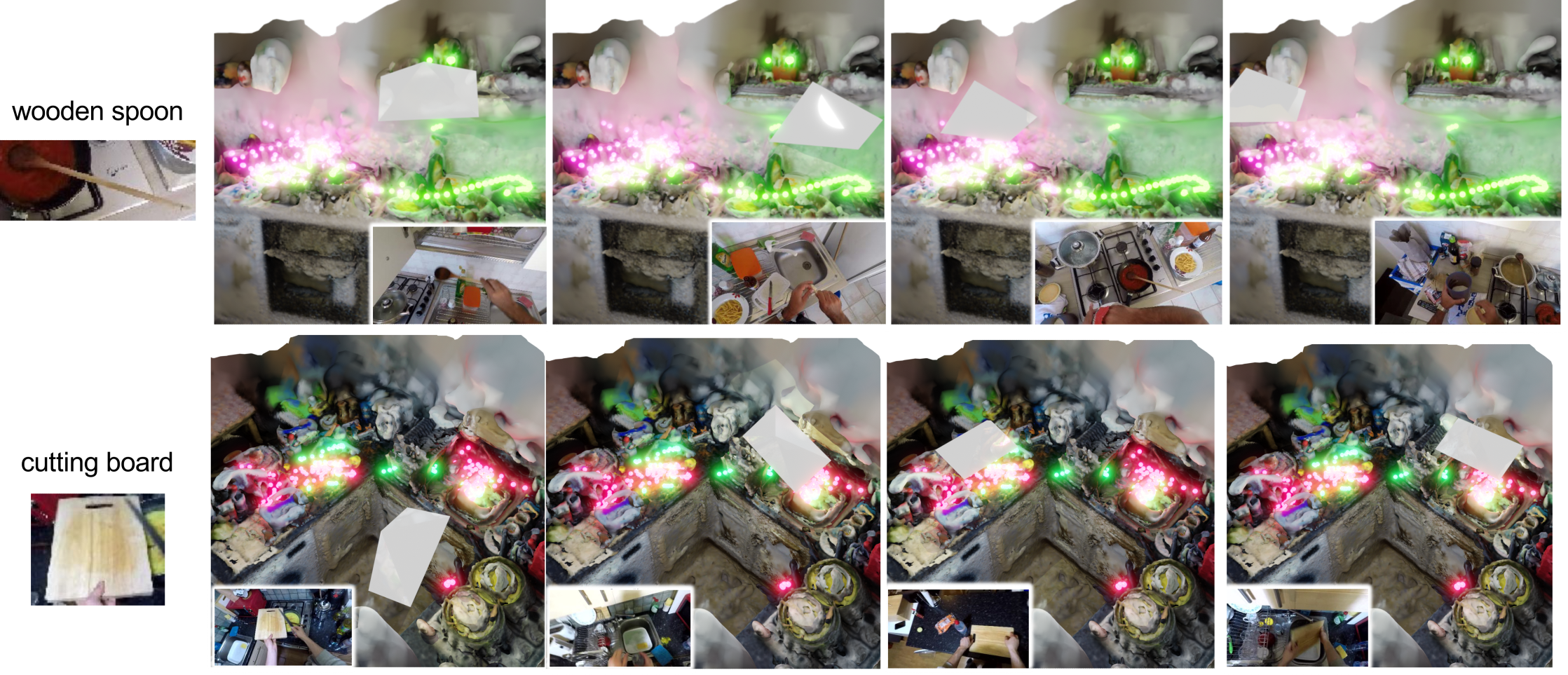}
    \caption{{\textbf{Trajectory prediction - lost track. }Predicted trajectory of two objects in motion. Green neon dots show correctly predicted 3D positions over four frames with their corresponding camera views, and red neon dots show ground-truth trajectory where the prediction fails. The tracking fails and all subsequent predictions are assigned to a new track. }}
    \label{fig:failure2}
\end{figure*}

We identify two key reasons for failure cases for LMK. For clarity, we showcase each case separately -- in Figure \ref{fig:failure1} and Figure \ref{fig:failure2}. For each figure, we focus on a single object and show its predicted trajectory in green. Failure predictions are shown in red, where we plot the correct ground truth trajectory.

In Figure \ref{fig:failure1} we show cases where the track is lost for a limited time but is then correctly recovered.
In the first row, the tin is correctly tracked for most of its trajectory, including when it is discarded in the bin. However, for a short duration, the predictions are incorrect (red dots).
Similarly, in the second row, the knife is incorrectly predicted while occluded by the hand or occluded in hand.
The last example shows failures in predicting the correct trajectory of the pot as it is filled with milk which changes its appearance. Coincidentally, it is moved out of the field of view. The matching then fails for both the appearance and the location. As the pot is emptied, its appearance matching is recovered towards the end of the track.

In Figure \ref{fig:failure2}, we show failure cases of tracking that are not recovered. 
In the first example, the wooden spoon is assigned a new trajectory and the tracking continues using the new identity. 
This is similarly the case for the cutting board when it is moved to the cluttered sink.

Failures predominantly occur in cluttered scenarios, such as when slicing peppers with a knife in Figure \ref{fig:failure1}, or mixing with a spoon in Figure \ref{fig:failure2}. In these situations, the locations of multiple objects overlap, making the individual object's location less informative for matching.

\section{Future Directions}

We report the majority of our results using ground-truth masks out of the VISOR annotations.
This allows us to evaluate the tracking  from partial observations without accumulating detection errors.
We find this decision to be reasonable as we focus on 
introducing and evaluating the task of Out of Sight, Not Out of Mind (OSNOM). 
In Fig~\ref{fig:dets}, we ablate this by using an off-the-shelf semantic-free detector. 
The figure shows an expected drop in performance as noisy and incomplete detections are introduced. 
Improving performance using detection predictions is one of the future directions.

Another future direction is the expansion of OSNOM task to multiple videos, over time. In follow-up videos, the initial assumption of where objects are from previous sessions can be used as priors for OSNOM. Extending beyond a single video targets our ultimate goal of an assistive solution that is aware of where objects are, over hours and potentially days.

\end{document}